\pdfoutput=1

\documentclass[11pt]{article}

\usepackage{emnlp2021}

\usepackage{times}
\usepackage{latexsym}

\usepackage[T1]{fontenc}

\usepackage[utf8]{inputenc}

\usepackage{microtype}

\usepackage{amsmath,amsfonts,bm}
\usepackage{booktabs} %
\usepackage{url}
\usepackage{graphicx}
\usepackage{array}
\usepackage{color}
\usepackage{makecell}
\usepackage{multirow}
\usepackage{xspace}
\usepackage{colortbl}

\newcommand{\ie}{\textit{i.e.}} %
\newcommand{\eg}{\textit{e.g.}} %
\newcommand{\start}[1]{\vspace{.3mm}\noindent{{\bf #1}.}}

\newcommand{\ours}{DDBA\xspace}
\newcommand{\framework}{EDE\xspace}
\newcommand{\task}{text-to-text generation\xspace}
\newcommand{\Task}{Text-to-text generation\xspace}

\newcommand{\upv}{\vspace{-.0cm}}
\newcommand{\downv}{\vspace{-.1cm}}

\definecolor{gred}{RGB}{219,68,55}
\definecolor{gblue}{RGB}{66,133,244}
\definecolor{gyellow}{RGB}{244,180,0}
\definecolor{ggreen}{RGB}{15,157,88}
\definecolor{ggrey}{RGB}{115,115,115}
\definecolor{na}{gray}{0.9}

\usepackage{amsmath,amsfonts,bm}

\def\eqref#1{equation~\ref{#1}}

\def\1{\bm{1}}

\DeclareMathAlphabet{\mathsfit}{\encodingdefault}{\sfdefault}{m}{sl}
\SetMathAlphabet{\mathsfit}{bold}{\encodingdefault}{\sfdefault}{bx}{n}

\title{Extract, Denoise and Enforce: Evaluating and Improving\\ Concept Preservation for Text-to-Text Generation}

\author{Yuning Mao$^{1}$, Wenchang Ma$^{2\star}$, Deren Lei$^{3\star}$, Jiawei Han$^{1}$, Xiang Ren$^3$ \\
$^1$University of Illinois at Urbana-Champaign \quad
$^2$Tsinghua University \quad
$^3$University of Southern California\\
$^1$\{yuningm2, hanj\}@illinois.edu $\quad$ $^2$mwc17@mails.tsinghua.edu.cn $\quad$ $^3$\{derenlei, xiangren\}@usc.edu
}

\begin{document}
\maketitle

{
\renewcommand{\thefootnote}{\fnsymbol{footnote}}
\footnotetext[1]{Equal contribution}
}

\begin{abstract}
Prior studies on \task typically assume that the model could figure out what to attend to in the input and what to include in the output via seq2seq learning, with only the parallel training data and no additional guidance.
However, it remains unclear whether current models can preserve important concepts in the source input, as seq2seq learning does not have explicit focus on the concepts and commonly used evaluation metrics also treat concepts equally important as other tokens.
In this paper, we present a systematic analysis that studies whether current seq2seq models, especially pre-trained language models, are good enough for preserving important input concepts and to what extent explicitly guiding generation with the concepts as lexical constraints is beneficial.
We answer the above questions by conducting extensive analytical experiments on four representative \task tasks.
Based on the observations, we then propose a simple yet effective framework to automatically extract, denoise, and enforce important input concepts as lexical constraints. This new method performs comparably or better than its unconstrained counterpart on automatic metrics, demonstrates higher coverage for concept preservation, and receives better ratings in the human evaluation.\footnote{Our code is available at \url{https://github.com/morningmoni/EDE}}
\end{abstract}

\section{Introduction}

\begin{figure}[t]
    \centering
    \includegraphics[width=.98\linewidth]{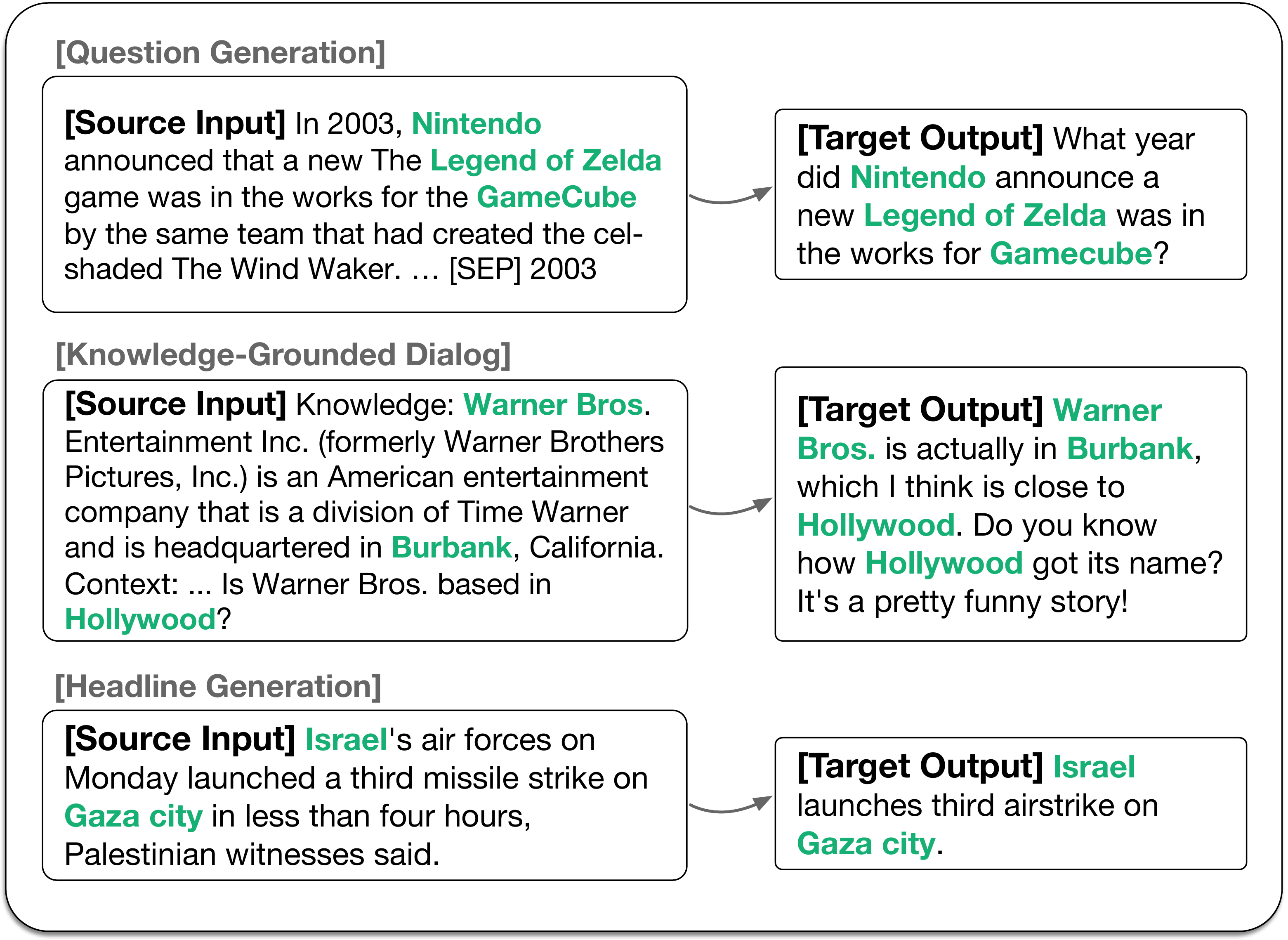}
    \upv
    \caption{Examples of \task tasks where preserving important input concepts is crucial for producing satisfactory results.}
    \label{fig_examples}
    \downv
\end{figure}

\Task is an important research problem with a broad set of applications, such as dialog response generation~\cite{dinan2018wizard}, headline generation \cite{gu2020generating}, and summarization \cite{mao-etal-2020-multi}.
A distinct feature of \task (vs. free-form text generation) is that it is often desired to preserve the concepts in the source input (see  Fig.~\ref{fig_examples} for an illustration).
On one hand, concept preservation is crucial for maintaining the factual consistency between the input and output \cite{maynez-etal-2020-faithfulness,nan-etal-2021-entity}.
On the other hand, encouraging the model to focus on important input concepts may also improve its generation quality \cite{yao2019plan,li2020keywords}.

Mainstream \task methods are mostly data-driven, which ``hope'' to learn meaningful mappings between source input and target output via sequence-to-sequence (seq2seq) learning -- this is particularly the case for the recent pre-trained language models (PLMs)~\cite{lewis-etal-2020-bart,raffel2020exploring}, where seq2seq learning is expected to identify \textit{what to attend to} in the source input and \textit{what to include} in the model output, with access to only parallel training data.

However, as seq2seq learning does not explicitly focus on key concepts (\eg, named entities) and commonly used evaluation metrics (\eg, BLEU and ROUGE) also treat all the tokens in a sequence equally important, it is unclear how many of the important input concepts can be (or have been) preserved.
Existing attempts to alleviate the above issue use soft constraints, such as copy mechanism or additional attention, to focus on (certain parts of) the source input \cite{see-etal-2017-get,dinan2018wizard,dou2020gsum}.
Nevertheless, they still lack explicit guidance and resort to seq2seq learning itself to figure out what is important, without any guarantee of the model output or evaluation on the input concept preservation.

Explicit guidance for text generation can be achieved by lexically (hard) constrained generation (LCGen), which specifies lexical constraints (tokens) that must be included in the model output.
However, to what extent guiding \task with lexical constraints works in general remains unknown, as existing studies on LCGen~\cite{hokamp-liu-2017-lexically,post-vilar-2018-fast,zhang-etal-2020-pointer} focus on scenarios where gold (ground-truth) constraints are given (\eg, generate a story using user-specified keywords), while in generic \task tasks the constraints (target output) are unavailable.

In this paper, we present a systematic analysis on generic \task to understand (1) the abilities of seq2seq models, especially the PLMs, for preserving important input concepts and (2) whether more explicit guidance that uses important input concepts as lexical constraints can complement seq2seq learning. 
We select four representative tasks where preserving important concepts (entities) in the source input and incorporating them in the model output is essential, including question generation, knowledge-grounded dialog, headline generation, and abstractive summarization.
We examine the effectiveness of guiding generation with the important input concepts as lexical constraints, where the concepts are either obtained by comparing the source input with the target output in an analytical study or automatically extracted from the source input in a practical setting.

Specifically, in the analytical study, we first evaluate how many of the concepts found in the target output (named \textit{gold concepts}) are available in the source input, and how many of them are already preserved by seq2seq models.
We then investigate the room for improvement if we guide generation with the gold concepts as lexical constraints (named \textit{gold constraints}).
Next, when the target output is unavailable, we propose a simple yet effective framework, named \framework, to automatically Extract, Denoise, and Enforce input concepts as lexical constraints. 
\framework achieves significant improvement on two tasks and moderately better or comparable performance on the other two under sequence-level automatic evaluation.
Moreover, \framework receives better ratings in the human evaluation and demonstrates higher coverage in the concept-level evaluation on all the tasks that we examine.

\start{Contributions}
(1) We analyze whether current PLMs for \task are good enough for preserving important input concepts via seq2seq learning.
(2) We study the usefulness of guiding generation explicitly with important input concepts as lexical constraints.
(3) We propose a framework to automatically extract, denoise, and enforce concept constraints, which achieves comparable or better performance than unconstrained generation on a range of \task tasks.

\section{Analysis on Concept Preservation}
\label{sec_simulated}

In this section, we conduct a series of analytic experiments on concept preservation, which tries to answer the following two questions:

\smallskip
\noindent
\textbf{Q1}: Is the current ``PLMs$+$seq2seq fine-tuning" paradigm for \task good at preserving the important concepts in the source input?

\smallskip
\noindent
\textbf{Q2}: What is the room for improvement if we guide \task by enforcing high-quality (gold) concepts as lexical constraints?

\subsection{Analysis Setup}

We conduct extensive analytical experiments on four \task tasks including question generation, knowledge-grounded dialog, headline generation, and abstractive summarization.
All the tasks require the model to preserve important concepts (entities) in the input and include them in the output in order to achieve satisfactory results.
Correspondingly, we consider the input as the source of lexical constraints and the entities in the target output as the desired (gold) constraints.

To answer Q1, we first evaluate the availability of concepts, namely how many of the gold concepts appearing in the target output can be found in the source input.
We then conduct a manual analysis to study why some of the gold concepts are missing in the source input.
Finally, we analyze model performance by matching gold concepts with the unconstrained model output, which reveals how many of the important input concepts are already preserved without the use of explicit constraints.

To answer Q2, we use the gold concepts as lexical constraints to guide the generation process.
We first estimate an ideal upper bound by enforcing all the gold constraints taken from the target output.
We next remove the gold constraints that cannot be found in the source input and examine how large the gap is from the ideal upper bound.

\subsection{Tasks and Datasets}

\start{Question Generation}
Question generation is the task of generating a question given a passage and the corresponding answer.
Ideally, a seq2seq model would learn to focus on the relevant information surrounding the answer span and reuse some of the concepts in the source input as part of the generated question.
We use the SQuAD 1.1 dataset \cite{rajpurkar-etal-2016-squad}, which is repurposed by \citet{du-etal-2017-learning} for question generation.

\start{Knowledge-Grounded Dialog}
Knowledge-grounded dialog involves utterances with groundings to specific knowledge sources.
It is used as another test case for our analysis as the model is supposed to extract important concepts from the grounded knowledge when appropriate.
We use the Wizard of Wikipedia (WoW) dataset \cite{dinan2018wizard} for evaluation.
As we focus on generation instead of knowledge retrieval, we adopt the gold knowledge setting of WoW where the sentences with relevant knowledge are provided and use the test split that consists of new dialogues under overlapping topics with the training set.

\start{Headline Generation}
Headline generation fits our analysis as a headline usually consists of the most important concepts in the input article.
We use the English Gigaword dataset \cite{napoles2012annotated} for evaluation.
As training on the full training set of Gigaword \cite{rush-etal-2015-neural} is computationally prohibitive, we adopt two low-resource settings where the 10k training examples in \citet{dong2019unified} and the first 300k of the 3.8M training examples are used.

\start{Abstractive Summarization}
Abstractive summarization is used as another testbed  as a summary generally involves important concepts in the source document.
We take two widely used news summarization datasets, CNN/Daily Mail (CNN/DM) \cite{nallapati-etal-2016-abstractive-new} and XSum \cite{narayan-etal-2018-dont}, for evaluation,
where CNN/DM is more extractive \cite{mao-etal-2020-facet} and XSum more abstractive \cite{maynez-etal-2020-faithfulness} in nature.

\subsection{Constrained Generation}
Next, we briefly introduce constrained generation and the specific methods used for our analysis.

\start{Hard Constrained Generation}
Lexically (hard) constrained generation (LCGen) specifies lexical constraints (tokens) that must be present in the model output.
Compared to soft constrained generation, LCGen has the advantage of ensuring the presence of certain input concepts explicitly, but it could also be problematic when the enforced constraints are noisy (inappropriate).
We choose from LCGen methods that enforce constraints by constraining beam search~\cite{hokamp-liu-2017-lexically,post-vilar-2018-fast}, as they function at the inference stage and can be easily combined with different seq2seq models, making our analysis more generalizable.
In contrast, sampling-based and insertion-based approaches are not easily applicable to generic \task, as they typically involve specialized training schemes or do not accept other inputs than the constraints \cite{miao2019cgmh,zhang-etal-2020-pointer,sha-2020-gradient}.

Specifically, we take dynamic beam allocation (DBA)~\cite{post-vilar-2018-fast} for our analysis, as it has higher efficiency than other LCGen methods that constrain beam search~\cite{hokamp-liu-2017-lexically}.
DBA revises the process of beam search by dividing the beam into a number of groups (named \textit{banks}), each of which stores the hypotheses that satisfy the same number of constraints. 
DBA can ensure that every constraint is present in the model output, as the <EOS> token is only allowed when all the constraints are met.
At each decoding step, there are three sources of candidates: (1) the top-k tokens across all hypotheses as in standard beam search; (2) all unfulfilled constraint tokens for each hypothesis; and (3) the single-best token for each hypothesis.
The banks are trimmed by the sequence probability if the total number of candidate tokens is beyond capacity (beam size).

\start{Soft Constrained Generation}
We additionally examine the effectiveness of soft constrained generation for comparison.
There are various types of soft constraints and we particularly consider the type which implicitly specifies input texts that the model needs to focus on.
We use copy mechanism as a representative example for this purpose.
Copy mechanism estimates the importance of tokens in the source input and learns to copy them when appropriate, which is useful for preserving the important concepts in the input, especially rare concepts that the model does not get enough exposure to during training.
For PLMs, we take the encoder-decoder cross attention in the last decoder layer as the copy distribution \cite{xu-etal-2020-self}.

\subsection{Experimental Settings}

We use BART \cite{lewis-etal-2020-bart} as the major base model for analysis.
We use spaCy~\cite{spacy2} to extract the entities from the target output as the gold concepts, the quality of which is shown to be reasonably good for the source-target alignment in summarization \cite{nan-etal-2021-entity}.
We mainly use exact matching to be consistent with the current automatic metrics that generally consider lexical overlap, while also manually analyzing the missing concepts to address the limitation of exact matching.
More implementation details can be found in App.~\ref{app_implementation}.

\subsection{Results and Analysis}

\start{Concept Availability}
We first examine how many of the gold concepts can be found in the source input.
A high degree of concept overlap between the source input and target output is expected, since the tasks require groundings to the input particularly.
As listed in the 3rd column of Table~\ref{tab_concept_stats}, even if the \textit{task} requires groundings to the input, it is not uncommon that the corresponding \textit{dataset} involves extrinsic information accidentally, \ie, the target output contains information that cannot be found in the source input. The issue is most severe on Gigaword and XSum: around half of the gold concepts are not found in the source input, which coincides with recent studies on the factual consistency of \task \cite{matsumaru-etal-2020-improving,maynez-etal-2020-faithfulness}.

\begin{table}[ht]
\centering
\resizebox{0.99\columnwidth}{!}{
\begin{tabular}{lcccc}  
\toprule[1.0pt]
\textbf{Task} & $|C|$ & $|C \cap X| / |C|$ & $|C \cap Y_{\text{sys}}| / |C|$ & $|C \cap Y_{\text{sys}}| / |C \cap X|$\\
\midrule[0.3pt]
SQuAD & 1.15 & 73.5\% & 46.6\% & 59.4\% \\
WoW & 1.01 & 70.4\% & 57.0\% & 78.9\% \\
Gigaword & 1.62 & \textbf{51.9\%} & 43.6\% & 75.4\%\\
CNN/DM & \textbf{6.68} & 88.7\% & 58.9\% & 66.0\% \\
XSum & 2.76 & \textbf{47.7\%} & 48.7\% & 69.7\% \\

\bottomrule[1.0pt]	
\end{tabular}
}
\upv
\caption{Statistics of concept availability (3rd column) and fulfillment (4th and 5th columns). $|C|$ denotes the number of gold concepts. $X$ and $Y_{\text{sys}}$ denote the concepts in source input and unconstrained model output.}
\downv
\label{tab_concept_stats}
\end{table}

\start{Manual Analysis on Missing Concepts}
To further study why certain gold concepts cannot be found in the source input, we conduct manual analysis of 50 examples on each dataset and categorize the reasons into the following groups:
(Spell): the target output and source input use different terms to refer to the same concept (\eg, synonyms, pronouns), which are not matched by lexical overlap.
(Miss): the concept is indeed missing in the input.
(NER): the concept is not a proper entity due to named entity recognition errors.
(Knowledge): the concept is rephrased with commonsense knowledge (\eg, ``on Valentine’s Day'' to ``in February'').

\begin{figure}[t]
    \centering
    \includegraphics[width=0.98\linewidth]{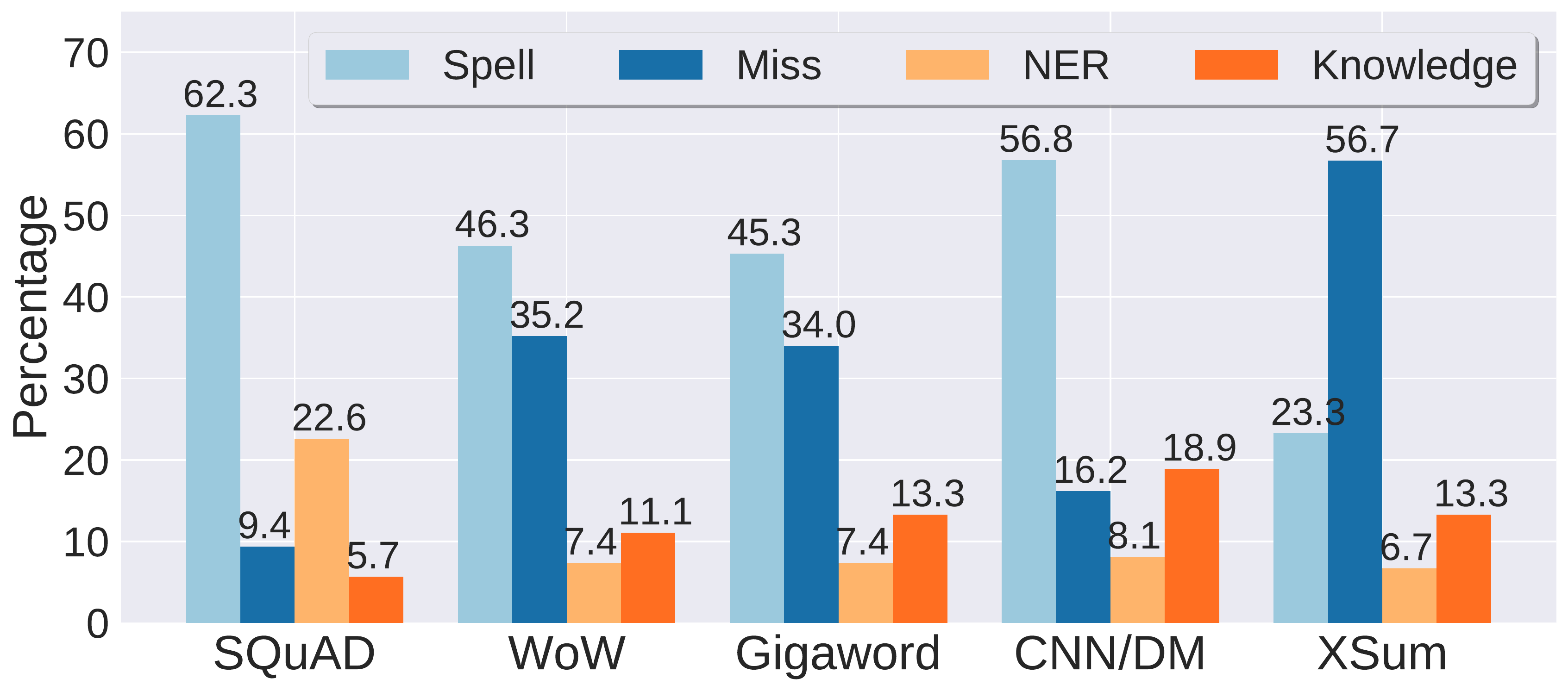}
    \upv
    \caption{Categories of why certain gold concepts in the target output cannot be found in the source input.}
    \label{fig_missing_bar}
    \downv
\end{figure}

As shown in Fig.~\ref{fig_missing_bar}, for most datasets, the major reason that gold concepts are missing is simply that they are spelled differently and not found by exact matching.
That said, three of five datasets still have 10\%+ missing gold concepts after ruling out this factor.\footnote{We obtain a percentage estimation of actual missing concepts by $(1 - |C \cap X| / |C|) \times \text{Miss}\%$, which is 2.5\%, 10.4\%, 16.4\%, 1.8\%, and 29.7\% for the 5 datasets, respectively.}
Such findings suggest that we need to put more effort to factual consistency when creating \task datasets and remove problematic (hallucinated) examples from existing datasets if possible.
The quality of gold concepts (NER) is reasonably high except for SQuAD. Nevertheless, after a closer look we find that most NER errors on SQuAD are repeated ones (\eg, ``what year'' is often incorrectly recognized as \textit{date}) that can be easily fixed manually.

\start{Concept Fulfillment}
We next study concept fulfillment, \ie, matching the gold concepts with the model output to examine how many of the important concepts have been preserved without the use of constraints.

As shown in the 4th column of Table~\ref{tab_concept_stats}, roughly half of the gold concepts are preserved through standard seq2seq learning.
The unconstrained model performs especially poorly on SQuAD and CNN/DM, as the gap between the concepts in the source input and those in the model output (the 3rd and 4th columns) is quite large.
Interestingly, the model output contains more gold concepts than the source input on XSum, possibly because the significant level of extrinsic information in the target output of XSum encourages the model to hallucinate.
If we only consider available gold concepts (the 5th column), the degree of concept fulfillment is improved consistently but remains relatively low.\footnote{The actual degree of concept fulfillment is likely to be higher as exact matching is used but the comparative analysis is still valid. More rigorously, one may again conduct manual analysis to categorize the missing concepts in model output.}

\begin{figure}[ht]
    \centering
    \includegraphics[width=0.98\linewidth]{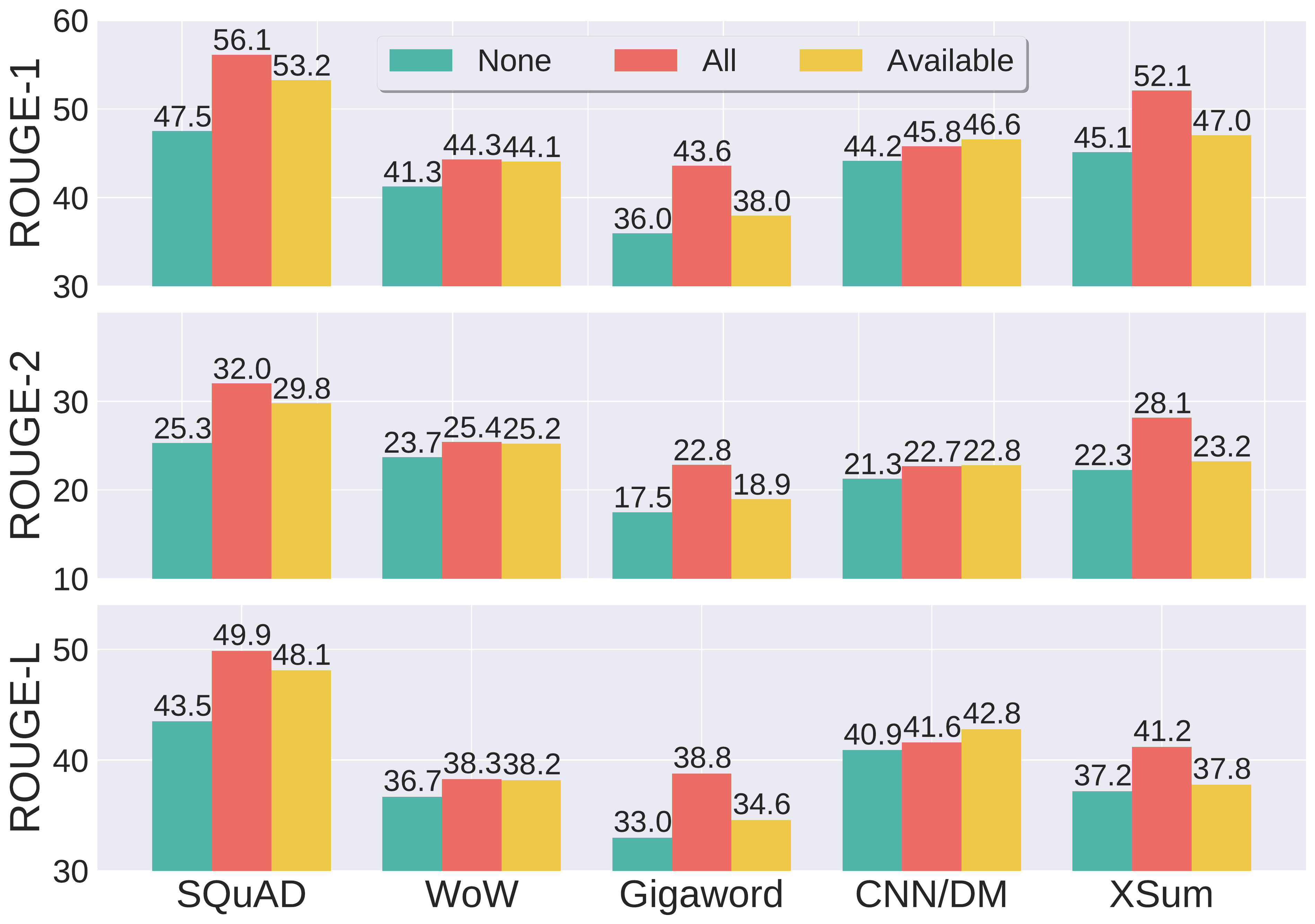}
    \upv
    \caption{Room for improvement of enforcing lexical constraints when all gold constraints (All) or only those found in the input are used (Available).}
    \label{fig_bound_bar}
    \downv
\end{figure}

\begin{figure*}[t]
    \centering
    \includegraphics[width=0.98\linewidth]{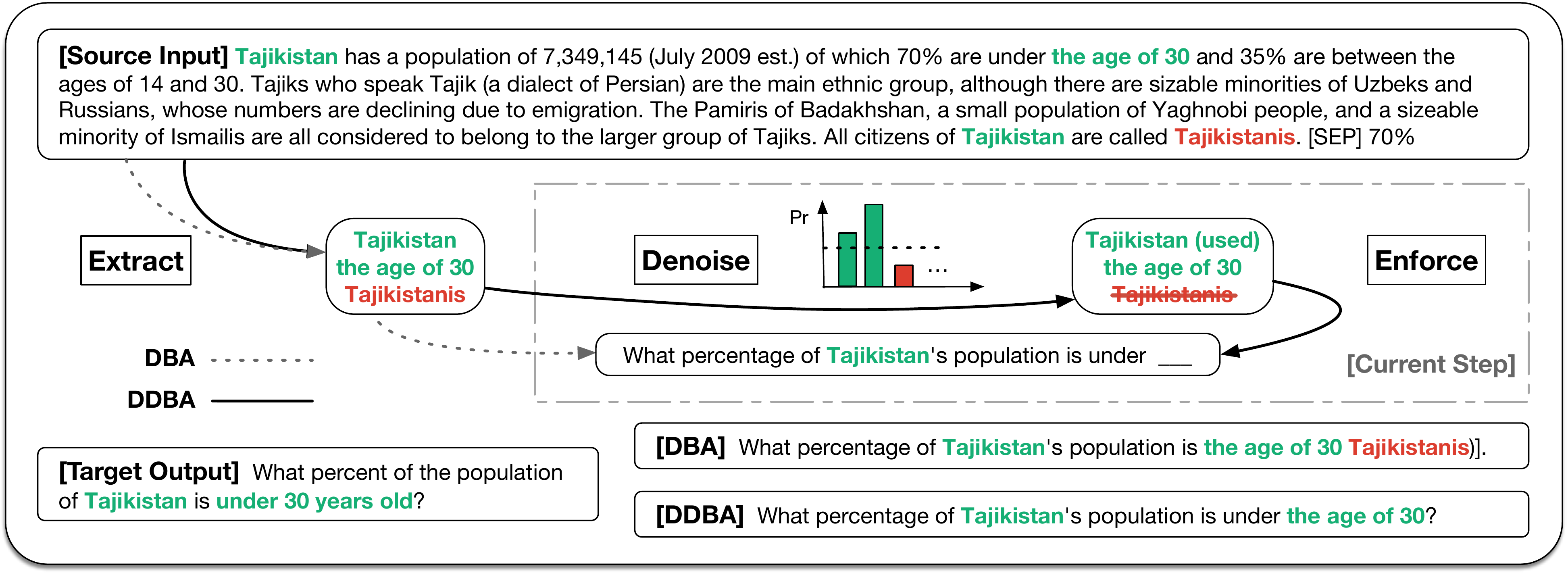}
    \upv
    \caption{\textbf{Illustration of the proposed framework \framework with a real example.} DBA enforces all lexical constraints and results in nonfluency, while \ours filters the noisy constraint and generates output with higher quality.}
    \label{fig_framework}
    \downv
\end{figure*}

\start{Upper bounding}
In Fig.~\ref{fig_bound_bar}, we list the ROUGE F1 scores when enforcing all the gold concepts as constraints (all) and only those found in the input (available).
We observe that the room for improvement is usually high when all the gold constraints are used, but may become smaller if those constraints not found in the source input are excluded, especially on Gigaword and XSum, the two datasets with more extrinsic information.
One exception is CNN/DM, where using available constraints performs better than using all.
We hypothesize that this is because CNN/DM involves longer output and much more concepts than other tasks, which together may mislead the beam search process if the beam is mostly occupied for fulfilling the constraints. A reduced number of constraints is hence more appropriate for it.

The results above indicate that when guiding generation with important input concepts as lexical constraints, it could be seemingly difficult to improve overall output quality on datasets that involve abundant extrinsic information.
That said, as commonly used metrics like ROUGE conduct sequence-level evaluation without distinguishing different tokens or identifying concepts, one also needs to examine model performance by measuring concept preservation in particular to truly reflect the effectiveness of LCGen (Sec.~\ref{sec_eval_concept}).

\section{Guiding Text-to-Text Generation with Extracted Concepts}
\label{sec_auto_constraint}

In this section, we study a practical setting where automatically extracted concepts are used as constraints (named \textit{automatic constraints}), with the following research question in mind:

\textbf{Q3}: Can we automatically extract important input concepts and use them as lexical constraints to guide \task?

\subsection{Automatic Constraint Extraction}
Existing studies on LCGen generally focus on scenarios where the gold constraints are provided and the model output is centered on the constraints, but for most applications in \task, the gold constraints (target output) are not accessible.
One thus needs to conduct automatic constraint extraction from the source input.
While it is not uncommon to extract keywords from the source input to help generation, previous methods \cite{yao2019plan,li2020keywords} typically use them as soft constraints via attention instead of lexical constraints that guide the generation process explicitly.

To extract constraints automatically, we create constraint labels on the training set by mapping gold concepts from the target output to the source input (as in the study of concept availability).
We then train a state-of-the-art keyphrase extraction model, BERT-KPE~\cite{sun2020joint}, to predict constraints during test time.
BERT-KPE uses contextualized representation and jointly optimizes keyphrase identification and ranking.
We conduct constraint extraction independently as our preliminary experiments suggest that multi-task learning of \task and constraint extraction does not appear particularly helpful. Also, decoupling the two tasks makes our framework generally applicable to different trained seq2seq models.

\subsection{Constraint Denoising}

Unlike using the gold constraints, directly enforcing all the automatic constraints is likely to worsen generation as some of the extracted constraints are inappropriate (see example in Fig.~\ref{fig_framework}).
Therefore, we propose a Denoised variant of DBA, named \ours, which is designed specifically for dealing with automatic constraints that are noisy in nature.
\ours conducts step-level dynamic constraint denoising by modifying the DBA method as follows.

First, only constraints deemed appropriate at a decoding step are added to the beam instead of all unmet constraints like DBA.
We use the generation probability in seq2seq models to measure the appropriateness of constraints.
Intuitively, inappropriate constraints that would cause nonfluency at the current step are likely to receive low probability scores and thus filtered.
Another advantage of dynamic constraint denoising is that by filtering noisy constraints, more beam space is saved for better sequences that are likely to lead to a final output with higher quality.
Second, unlike DBA, \ours allows the <EOS> token even when not all the constraints are fulfilled. In this way, the model is not forced to include noisy constraints in its output. 
Note that not satisfying all constraints is acceptable since our goal is not to fulfill \textit{all} the constraints but rather use them to guide generation.

\start{Comparison w. Supervised Denoising}
We considered training supervised classifiers using features such as self-attention and copy distribution to predict constraint quality, but found that simply filtering the constraints by their token probability at the current decoding step performs competitively with better efficiency, and hence use the vocabulary probability distribution as the scoring function instead (more details are in App.~\ref{sec_supervised_denoising}).
We observe that \ours, despite its simplicity, improves the quality, especially fluency, of constrained generation under both automatic and human evaluations.

\section{Experiments}
\label{sec:extrinsic}

In this section, we conduct experiments of guiding generation with automatic constraints and compare with the base as well as state-of-the-art models.
The constraints are automatically extracted \textit{without access to the target output}.

\subsection{Automatic Evaluation}
\start{Question Generation}
We show the performance comparison of question generation on SQuAD in Table~\ref{tab_QG-SOTA}.
\framework (DBA) leads to worse BLEU-4 and ROUGE-L but higher METEOR, which is possibly due to the noise in the automatic constraints and the fact that METEOR uses stemming and synonymy matching in addition to exact matching, which successfully matches terms with different spellings.
In contrast, \framework (\ours) performs significantly better than \framework (DBA) and vanilla unconstrained BART (up to +4.3/+2.5 on BLEU-4), consistently outperforming existing baselines with BART$_{\text{base}}$ and \textit{achieves new state-of-the-art results} with BART$_{\text{large}}$ on BLEU-4 and ROUGE-L.
Copy mechanism generally leads to higher ROUGE-L but worse performance on the other two metrics.

\begin{table}[ht]
\centering
\resizebox{0.99\columnwidth}{!}{
\begin{tabular}{lccc}  
\toprule[1.0pt]
\textbf{Method} & \textbf{{ BLEU-4}}~ & \textbf{{ METEOR}} & \textbf{{ R-L}}  \\
\midrule[0.3pt]
NQG \cite{du-etal-2017-learning}&12.28& 16.62& 39.75\\
SemQG \cite{zhang-bansal-2019-addressing}&20.76&24.20&48.91\\
ERNIE-G{\footnotesize EN}$_{\text{base}}$ \cite{xiao2020ernie}\ \ &22.28&25.13&50.58\\
BART$_{\text{base}}$ & 22.46&	24.29&	49.65\\
\cellcolor{na}\framework (BART$_{\text{base}}$+DBA) & 20.37& 25.30& 47.42	\\
\cellcolor{na}\framework (BART$_{\text{base}}$+\ours) & \textbf{24.57} & \textbf{25.82} & \underline{51.73}\\
BART-COPY$_{\text{base}}$ & 22.36&	24.21&	50.01\\
\cellcolor{na}\framework (BART-COPY$_{\text{base}}$+DBA)& 20.27&	25.17&	48.27\\
\cellcolor{na}\framework (BART-COPY$_{\text{base}}$+\ours) & \underline{24.02}&	\underline{25.59}&	\textbf{52.20}\\

\midrule[0.3pt]
UniLM$_{\text{large}}$ \cite{dong2019unified}&22.12&25.06&51.07\\
ERNIE-G{\footnotesize EN}$_{\text{large}}$ \cite{xiao2020ernie}&\underline{25.40}&\textbf{26.92}&\underline{52.84}\\
BART$_{\text{large}}$ & 23.22&	24.89&	50.03\\
\cellcolor{na}\framework (BART$_{\text{large}}$+DBA) & 21.40& 25.69& 47.99	\\
\cellcolor{na}\framework (BART$_{\text{large}}$+\ours) & \textbf{25.77}&	\underline{26.77}&	52.68\\
BART-COPY$_{\text{large}}$  & 22.95&	24.98&	50.90\\
\cellcolor{na}\framework (BART-COPY$_{\text{large}}$+DBA)& 20.81 & 25.78 & 49.18\\
\cellcolor{na}\framework (BART-COPY$_{\text{large}}$+\ours) & 25.00&	26.61&	\textbf{53.29}\\

\bottomrule[1.0pt]	
\end{tabular}
}
\upv
\caption{Performance comparison of question generation on SQuAD. R is short for ROUGE. \textbf{Best} and \underline{2nd best} methods are bold and underlined, respectively.}
\label{tab_QG-SOTA}
\downv
\end{table}

\start{Knowledge-Grounded Dialog}
We show the results of knowledge-grounded dialog on WoW in Table~\ref{tab_dialog-SOTA}.
We observe that adding both hard and soft constraints helps with the generation process, which again implies that more explicit guidance than pure seq2seq learning could still be beneficial even for the PLMs. 
The improvement is consistent with the relatively high upper bound of WoW when available gold constraints are used.
Also, enforcing all constraints via DBA without dynamic filtering seems to be more detrimental than helpful.

\begin{table}[t]
\centering
\resizebox{0.99\columnwidth}{!}{
\begin{tabular}{lcccc}  
\toprule[1.0pt]
\textbf{Method} & \textbf{F1} & \textbf{R-1} & \textbf{R-2} & \textbf{R-L} \\

\midrule[0.3pt]
ESCA-BERT \cite{haonan2020exploring} &29.1	&30.35	&16.43	&28.07\\
Two-Stage MemNet \cite{dinan2018wizard} &30.7 & 31.04 & 16.89 & 28.72 \\
E2E MemNet \cite{dinan2018wizard}&35.5& 37.34& 19.76 & 33.02\\

\midrule[0.3pt]

BART$_{\text{base}}$ & 39.1&	40.63&	22.93 & 35.82\\
\cellcolor{na}\framework (BART$_{\text{base}}$+DBA) & 37.7 &39.22	 &20.99	 &34.20  \\
\cellcolor{na}\framework (BART$_{\text{base}}$+\ours) & \underline{39.5} & \underline{41.25} & \textbf{23.69} & \textbf{36.77} \\
BART-COPY$_{\text{base}}$  & 39.4&	40.99&	23.45 & 36.39\\
\cellcolor{na}\framework (BART-COPY$_{\text{base}}$+DBA) & 38.8	&40.35	&22.51	&35.35\\
\cellcolor{na}\framework (BART-COPY$_{\text{base}}$+\ours) & \textbf{39.7}&	\textbf{41.37}&	\underline{23.66} & \underline{36.76}\\

\bottomrule[1.0pt]	
\end{tabular}
}
\upv
\caption{Performance comparison of knowledge-grounded dialog on WoW. R is short for ROUGE.}
\label{tab_dialog-SOTA}
\downv
\end{table}

\start{Headline Generation}
We show the results of headline generation in Table~\ref{tab_HG-SOTA}.
The improvement of \framework on Gigaword is not as much as previous tasks, which is probably because the concept availability on Gigaword is low and it is hard to extract and utilize meaningful concepts from the source input.
Nevertheless, we will later show that \framework is preferred in the human evaluation and also preserves the input concepts better on Gigaword.

\begin{table}[h]

\centering
\resizebox{0.99\columnwidth}{!}{
\begin{tabular}{clccc}  
\toprule[1.0pt]
&\textbf{Method} & \textbf{R-1} & \textbf{R-2} & \textbf{R-L} \\

\midrule[0.3pt]
\multirow{9}{*}{ \rotatebox[origin=c]{90}{10k training}}
&Transformer \cite{vaswani2017attention} & 10.97& 2.23& 10.42\\
&MASS \cite{song2019mass}& 25.03& 9.48& 23.48\\
&UniLM$_{\text{large}}$ \cite{dong2019unified}&32.96& 14.68& 30.56\\
&BART$_{\text{large}}$ & 34.08 & \underline{15.45} & 31.28\\
&\cellcolor{na}\framework (BART$_{\text{large}}$+DBA) & 33.12 & 14.79 & 30.24\\
&\cellcolor{na}\framework (BART$_{\text{large}}$+\ours) & \underline{34.15} & \textbf{15.56} & \underline{31.29}\\
&BART-COPY$_{\text{large}}$  & 34.10 & 15.09 & 31.24\\
&\cellcolor{na}\framework (BART-COPY$_{\text{large}}$+DBA) & 33.06 & 14.20 & 30.00\\
&\cellcolor{na}\framework (BART-COPY$_{\text{large}}$+\ours) & \textbf{34.22} & 15.23 & \textbf{31.36}\\

\midrule[0.3pt]
\multirow{6}{*}{ \rotatebox[origin=c]{90}{300k training}}
&BART$_{\text{base}}$ & 35.98&	17.50& 33.07\\
&\cellcolor{na}\framework (BART$_{\text{base}}$+DBA) & 34.77&	16.08&	31.30\\
&\cellcolor{na}\framework (BART$_{\text{base}}$+\ours) & 36.04& 17.52&  33.09\\
&BART-COPY$_{\text{base}}$  & \underline{36.35}&	\textbf{17.78}&	\underline{33.55}\\
&\cellcolor{na}\framework (BART-COPY$_{\text{base}}$+\ours) & 35.50 & 15.58 & 32.43\\
&\cellcolor{na}\framework(BART-COPY$_{\text{base}}$+\ours) & \textbf{36.39}&	\underline{17.72}&	\textbf{33.78}\\

\bottomrule[1.0pt]	
\end{tabular}
}
\upv
\caption{Performance comparison of headline generation on Gigaword. We only compare with BART when using 300k training examples as state-of-the-art methods use the full training set.}
\label{tab_HG-SOTA}
\downv
\end{table}

\start{Abstractive Summarization}
We list the results of abstractive summarization in Table~\ref{tab_summ-SOTA}.
The performance of \framework is comparable to its unconstrained counterpart on both datasets. Better performance may be achieved when the constraint extraction method is improved.
Apart from sequence-level evaluation (ROUGE), we observe that \framework is consistently better at concept preservation in concept-level evaluation (Sec.~\ref{sec_eval_concept}).

\begin{table}[t]
\centering
\resizebox{0.99\columnwidth}{!}{
\begin{tabular}{clccc}  
\toprule[1.0pt]
&\textbf{Method} & \textbf{R-1} & \textbf{R-2} & \textbf{R-L} \\
\midrule[0.7pt]

\multirow{5}{*}{ \rotatebox[origin=c]{90}{CNN/DM}}
&BertSum \cite{liu-lapata-2019-text}&42.13	&19.60	&39.18\\
&PEGASUS \cite{zhang2019pegasus}	&44.17	&21.47	&41.11\\

\cmidrule[0.3pt]{2-5}
&BART$_{\text{large}}$  & \textbf{44.16}	&\textbf{21.28}	&\textbf{40.90}\\
&\cellcolor{na}\framework (BART$_{\text{large}}$+DBA) & 43.22	&20.37	&40.04\\
&\cellcolor{na}\framework (BART$_{\text{large}}$+\ours) & \underline{44.06}	&\underline{20.41}	&\underline{40.89}\\

\midrule[0.7pt]
\multirow{5}{*}{ \rotatebox[origin=c]{90}{XSum}} &
BertSum \cite{liu-lapata-2019-text}&38.81	&16.50	&31.27\\
&PEGASUS {\cite{zhang2019pegasus}}	&47.21	&24.56	&39.25\\

\cmidrule[0.3pt]{2-5}
&BART$_{\text{large}}$  & 45.14	&22.27	&\textbf{37.25}\\
&\cellcolor{na}\framework (BART$_{\text{large}}$+DBA) & \underline{45.33}	&\underline{22.32}	&37.12\\
&\cellcolor{na}\framework (BART$_{\text{large}}$+\ours) & \textbf{45.40}	&\textbf{22.38}	&\underline{37.18}\\

\bottomrule[1.0pt]	
\end{tabular}
}
\upv
\caption{Performance comparison of abstractive summarization on CNN/DM and XSum.}
\label{tab_summ-SOTA}
\downv
\end{table}

\subsection{Human Evaluation}
In addition to automatic evaluation, we further conduct human evaluation with the following three aspects: closeness, relevancy, and fluency \cite{prabhumoye-etal-2019-towards}.
Closeness measures the similarity between the model output and target output.
Relevancy considers the quality of model output directly with the source input, since there are usually more valid outputs than the target output for generation tasks.
Fluency of the model output is on a scale of 1 (unreadable) to
4 (perfect).

We randomly sample 50 examples on each task and conduct pairwise comparisons between BART, \framework (DBA), and \framework (\ours) with the help of three external annotators.
As listed in Table~\ref{tab_human_eval}, the results are largely consistent with automatic evaluation.
For example, DBA leads to a similar number of better and worse outputs over BART on SQuAD and more worse cases on Gigaword, while \ours outperforms BART more stably on both datasets for closeness as well as relevancy.
The gaps are generally larger on relevancy than closeness, indicating that \ours is preferred by humans when not comparing with the target output directly. 
\ours also consistently outperforms DBA on both aspects and largely shares more similarities (ties) with BART due to its denoising function.
The fluency ratings of BART, DBA, and \ours are 3.82, 3.52, and 3.88 on SQuAD, and 3.34, 2.92, 3.26 on Gigaword. 
Such results indicate that DBA has negative impacts on generation fluency, while the fluency of \ours is comparable and sometimes even better than unconstrained generation.

\begin{table}[h]
\centering
\resizebox{0.95\columnwidth}{!}{
\begin{tabular}{llccc}  
\toprule[1.0pt]
\textbf{Task} & \textbf{Model} & \textbf{Win} & \textbf{Tie} & \textbf{Lose} \\

\midrule[0.6pt]
\fbox{SQuAD} & DBA vs. BART &24\% &50\%  &26\% \\
Closeness& \ours vs. BART &\textbf{16\%} &76\% &8\%\\
& \ours vs. DBA &\textbf{30\%}  &46\% &24\%\\
\midrule[0.3pt]
 & DBA vs. BART &\textbf{18\%} &66\% &16\%\\
Relevancy& \ours vs. BART &\textbf{16\% }&82\% &2\%\\
& \ours vs. DBA &\textbf{24\%} &68\% &8\%\\

\midrule[0.6pt]
\fbox{Gigaword} & DBA vs. BART &14\% &64\%  &\textbf{22\%} \\
Closeness& \ours vs. BART & \textbf{18\%} &68\% &14\%\\
& \ours vs. DBA & \textbf{24\%}  &72\% &4\%\\
\midrule[0.3pt]
 & DBA vs. BART &22\% &50\% &\textbf{28\%}\\
Relevancy& \ours vs. BART &\textbf{26\%} &54\% &20\%\\
& \ours vs. DBA & \textbf{26\%} &70\% &4\%\\

\bottomrule[1.0pt]	
\end{tabular}
}
\upv
\caption{Pairwise human evaluation comparing BART, \framework (DBA), and \framework (\ours). We list the results on two datasets here and provide the rest in App.~\ref{sec_human_eval}.}
\label{tab_human_eval}
\downv
\end{table}

\subsection{Evaluation on Concept Preservation}
\label{sec_eval_concept}
As sequence-level metrics consider all the tokens equally important and can only measure overall generation quality, we further conduct concept-level evaluation to measure model performance on concept preservation in particular.
We first examine the quality of extracted concepts (automatic constraints) in Table~\ref{tab_constraint_quality}. 
We observe that the F1 scores on different datasets are largely in the range of 0.4 to 0.5, which is on par with the state-of-the-art performance on keyphrase extraction benchmarks \cite{meng-etal-2017-deep}.

\begin{table}[h]
\centering
\resizebox{0.72\columnwidth}{!}{
\begin{tabular}{lccc}  
\toprule[1.0pt]
\textbf{Task} & \textbf{Precision} & \textbf{Recall} & \textbf{F1} \\
\midrule[0.3pt]
SQuAD & 29.1 & 62.9 & 39.8 \\
WoW & 35.4 &	53.1&	42.5\\
Gigaword & 41.1	&47.8	&44.2\\
CNN/DM &52.0	&50.1	&51.0\\
XSum & 38.5	&58.3	&46.4\\

\bottomrule[1.0pt]	
\end{tabular}
}
\upv
\caption{Quality of automatically extracted concepts (constraints) on different \task tasks. }
\label{tab_constraint_quality}
\downv
\end{table}

\begin{figure}[t]
    \centering
    \includegraphics[width=0.98\linewidth]{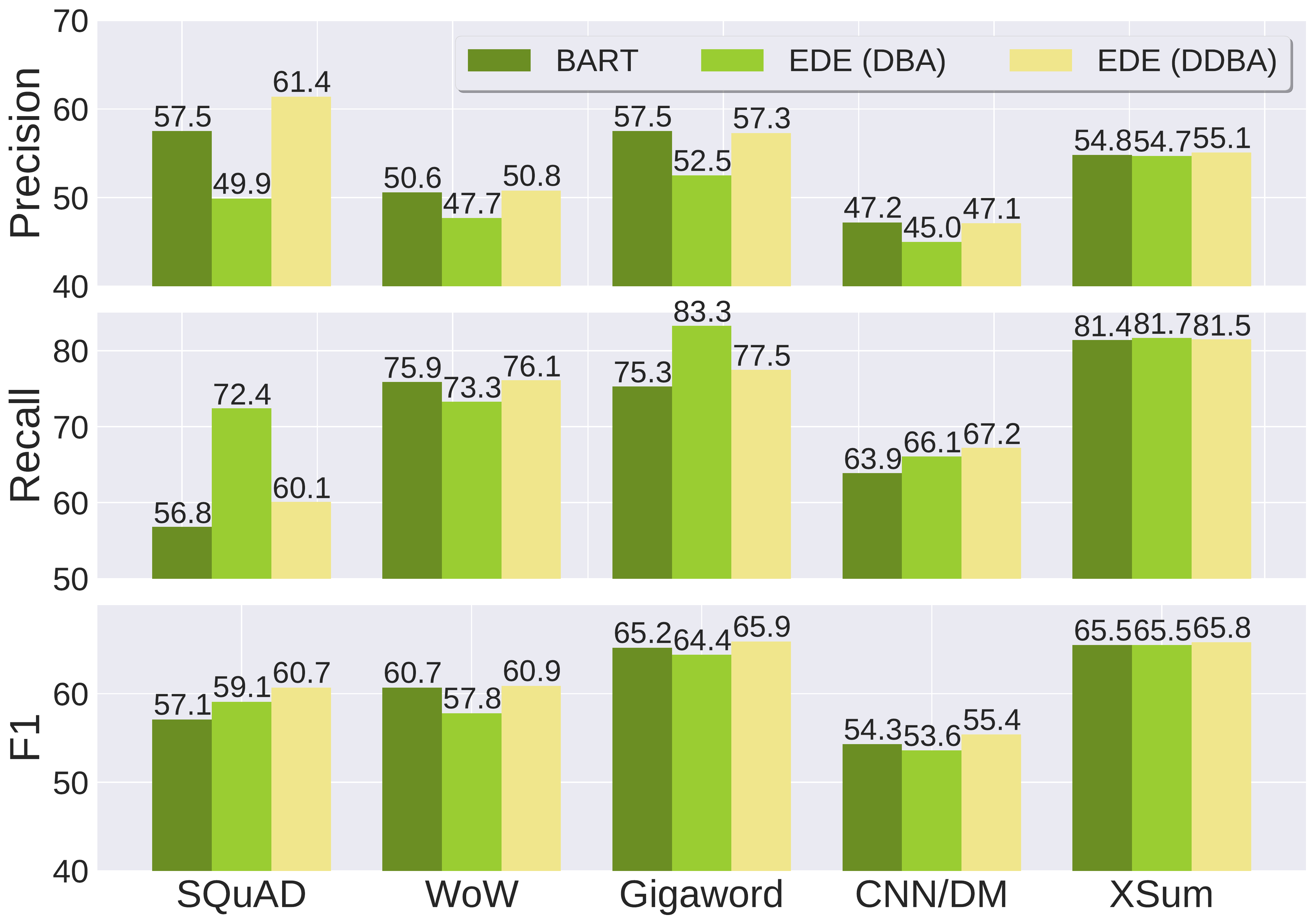}
    \upv
    \caption{Comparison on concept preservation. \framework (\ours) consistently achieves the best F1 score across different tasks. }
    \label{fig_concept_bar}
    \downv
\end{figure}

Similarly, we next analyze concept preservation of seq2seq models and show the results in Fig.~\ref{fig_concept_bar}.
We only consider gold concepts available in the source input for a fair evaluation.
We observe that DBA generally leads to higher recall but lower precision, while \ours balances the two metrics and consistently achieves the best F1.
The improvements are most remarkable on datasets where the unconstrained model could not preserve input concepts well (\eg, +3.6 on SQuAD and +1.1 on CNN/DM) and less significant if the unconstrained model already preserves most of the concepts or the dataset involves abundant extrinsic information.

\subsection{Takeaways}

From the analysis on concept preservation and model performance when automatic constraints are used, we can see that \framework is likely to improve \textit{overall} generation quality when the upper bound performance using available gold concepts is high.
\framework may not be very effective when many gold concepts cannot be found from the source input, which is an issue caused by mixed factors: (Dataset) the target output itself could be problematic and involve extrinsic concepts; (Model) concept extraction uses exact matching and misses some of the concepts; (Evaluation) existing metrics also use exact matching and would not recognize model improvement even if the concepts with different spellings are extracted.
That said, \framework still achieves comparable or better performance than its unconstrained counterpart under sequence-level evaluation (overall quality) and better preserves input concepts under concept-level evaluation (concept preservation).

\section{Related Work}

\start{Hard Constrained Generation}
Lexically (hard) constrained generation (LCGen) has been adopted in various applications.
One line of LCGen methods involves specialized model design or training schemes, which is thus not easily applicable to generic \task \cite{miao2019cgmh,zhang-etal-2020-pointer,sha-2020-gradient}. Other methods constrain the search space during decoding and can be plugged into fine-tuned models without additional training \cite{hokamp-liu-2017-lexically,post-vilar-2018-fast}.
The constraints used in existing studies are either from external sources \cite{hokamp-liu-2017-lexically}  or taken directly from the target output \cite{post-vilar-2018-fast,zhang-etal-2020-pointer,sha-2020-gradient}.
Different from the previous settings of LCGen where the goal is to generate outputs satisfying the specified constraints, our study focuses on how to use lexical constraints to guide generation in more generic tasks. That is, the model can decide to take (or ignore) any constraints as long as the decisions are beneficial for generation. 

\start{Soft Constrained Generation}
Soft constrained generation does not specify lexical constraints explicitly but encourages the model to attend to certain input texts.
In \task tasks where incorporating important concepts of the source input into the model output is essential, the soft constraints are usually achieved by attention mechanism to the source document \cite{see-etal-2017-get,dou2020gsum} or some additional inputs such as keywords \cite{yao2019plan,li2020keywords} and external knowledge~\cite{dinan2018wizard}.

\start{Factual Consistency of Text-to-Text Generation}
Accurately preserving the important concepts of the source input is critical for many \task tasks to ensure the factual consistency between the source input and model output.
However, recent studies \cite{maynez-etal-2020-faithfulness,zhou2020detecting} show that models learned in a seq2seq manner are prone to hallucinate unfaithful or nonfactual information, hindering their applicability to real-world applications.
Guiding \task with explicit constraints has been recently shown to help alleviate model hallucination and improve factual consistency \cite{mao2020constrained}.

\section{Conclusion}
In this work, we examine whether current pre-trained language models for \task are good enough for preserving important concepts in the source input without explicit guidance but pure seq2seq learning.
We conduct extensive analytical experiments on a range of \task tasks and study when adding important input concepts as lexical constraints can help guide \task.
We propose a simple yet effective framework for automatic constraint extraction, denoising, and enforcement, which is shown to perform comparably or better than unconstrained generation in various \task tasks and better preserves important input concepts.

\section*{Acknowledgements}
We thank Yiqing Xie and anonymous reviewers for helpful feedback.
Research was supported in part by US DARPA KAIROS Program No. FA8750-19-2-1004 and SocialSim Program No.  W911NF-17-C-0099, National Science Foundation IIS-19-56151, IIS-17-41317, and IIS 17-04532, and the Molecule Maker Lab Institute: an AI Research Institutes program supported by NSF under Award No. 2019897. Any opinions, findings, and conclusions or recommendations expressed herein are those of the authors and do not necessarily represent the views, either expressed or implied, of DARPA or the U.S. Government.

\bibliography{anthology,custom}
\bibliographystyle{acl_natbib}

\clearpage
\appendix

\section{Implementation Details}
\label{app_implementation}
We use Nvidia GTX 2080 Ti and Quadro RTX 8000 for the training of BART$_{\text{base}}$ and BART$_{\text{large}}$, respectively.
The evaluation metrics for each task are consistent with existing methods for comparison.
For constraint extraction, we remove a constraint completely before entering the generation process if its score predicted by the keyphrase extraction model is lower than a threshold (tuned to optimize constraint F1 score for each task).
LCGen is typically slower than standard beam search but the runtime overhead is acceptable when the beam size is not too large (we set it no larger than 20) and the number of constraints is small (most tasks involve one to two constraints on average).

\section{Supervised Denoising}
\label{sec_supervised_denoising}
In our preliminary experiments, we considered training a supervised classifier to estimate step-level constraint importance during decoding.
Specifically, we use the following features to measure constraint importance: two dynamic features based on the token probability in the vocabulary distribution and the token probability in the copy distribution (if available); Two static features, out-degree and in-degree centrality defined below, based on the transformer self-attention of source tokens.

Source token centrality is based on a directed graph built upon the self-attention \cite{xu-etal-2020-self}. Let $G=(V, E)$ denote a graph representing the encoder self-attention, where vertices $V$ denote the source tokens and $E(i, j)$ denote the self-attention from token $i$ to $j$ with $\sum_i E(i, j) = 1$.
The out-degree centrality of token $i$ is defined as $\sum_j E(i, j)$, which measures the degree a token $i$ contributes to other tokens. 
A transition probability matrix $T$ is defined as $T(i, j) = E(i, j) / \sum_j E(i, j)$, where the self-attention is normalized by the out-degree centrality.
The in-degree centrality of token $i$ is defined as $\sum_j T(j, i)$.

We first run \framework and store the intermediate dynamic features during decoding.
Then, those gold constraints with a generation probability greater than a threshold at a certain decoding step are treated as positive examples. As the features are all scalar, we use random forest as the classifier due to efficiency considerations. Other conventional classifiers performed worse in our experiments.
The comparison between \ours (supervised) and \ours (unsupervised) is shown in Table~\ref{tab_supervised}.
We observe that the performance of the two variants is rather similar, which is possibly because there are no step-level labels that determine whether it is appropriate to use a constraint at a specific decoding step, and we have to use approximate labels that also depend on the generation probability.

We also explored using reinforcement learning that assigns sequence-level reward (\eg, the scores on evaluation metrics) to bypass the lack of step-level labels. However, reinforcement learning turned out to be unstable and costly to learn.

\begin{table}[ht]
\centering
\resizebox{0.99\columnwidth}{!}{
\begin{tabular}{lccc}  
\toprule[1.0pt]
\textbf{Method} & \textbf{{ BLEU-4}}~ & \textbf{{ METEOR}} & \textbf{{ R-L}}  \\
\midrule[0.3pt]
\ours (supervised) &24.14 &25.43 & 51.67\\
\ours (unsupervised) & 23.98&	25.44&	51.69\\

\bottomrule[1.0pt]	
\end{tabular}
}
\upv
\caption{Comparison of \ours (supervised) and \ours (unsupervised) on question generation.}
\label{tab_supervised}
\downv
\end{table}

\section{Human Evaluation}
\label{sec_human_eval}

In Table~\ref{tab_human_eval2}, we provide additional results on human evaluation.
The results are largely consistent with automatic evaluation: \framework (\ours) performs significantly better than BART on question and dialog generation, moderately better on headline generation, and comparable on summarization.

\begin{table}[h]
\centering
\resizebox{0.95\columnwidth}{!}{
\begin{tabular}{llccc}  
\toprule[1.0pt]
\textbf{Task} & \textbf{Model} & \textbf{Win} & \textbf{Tie} & \textbf{Lose} \\

\midrule[0.6pt]
\fbox{WoW} & DBA vs. BART &12\% &60\%  &28\% \\
Closeness& \ours vs. BART &18\% &70\% &12\%\\
& \ours vs. DBA &32\%  &52\% &16\%\\
\midrule[0.3pt]
 & DBA vs. BART &14\% &62\% &24\%\\
Relevancy& \ours vs. BART &22\% &64\% &14\%\\
& \ours vs. DBA &34\% &54\% &12\%\\

\midrule[0.6pt]
\fbox{CNN/DM} & DBA vs. BART &16\% &58\%  &26\% \\
Closeness& \ours vs. BART & 12\% &78\% &10\%\\
& \ours vs. DBA & 24\%  &70\% &6\%\\
\midrule[0.3pt]
 & DBA vs. BART &26\% &50\% &24\%\\
Relevancy& \ours vs. BART &16\% &70\% &14\%\\
& \ours vs. DBA & 20\% &68\% &12\%\\

\midrule[0.6pt]
\fbox{XSum} & DBA vs. BART &14\% &78\%  &8\% \\
Closeness& \ours vs. BART & 16\% &78\% &6\%\\
& \ours vs. DBA & 20\%  &70\% &10\%\\
\midrule[0.3pt]
 & DBA vs. BART & 10\% &78\% &12\%\\
Relevancy& \ours vs. BART & 16\% &70\% &14\%\\
& \ours vs. DBA & 16\% &76\% &8\%\\

\bottomrule[1.0pt]	
\end{tabular}
}
\upv
\caption{Pairwise human evaluation comparing the base model BART, \framework (DBA), and \framework (\ours).}
\label{tab_human_eval2}
\downv
\end{table}

\end{document}